\pdfoutput=1

\documentclass[11pt]{article}

\usepackage{acl}

\usepackage{times}
\usepackage{latexsym}
\usepackage{graphicx}
\usepackage{booktabs}
\usepackage{amsmath}
\usepackage{bbm}
\usepackage{subfigure}

\usepackage[T1]{fontenc}

\usepackage[utf8]{inputenc}

\usepackage{microtype}

\title{Continuous Temporal Graph Networks for Event-Based Graph Data}

\author{Jin Guo$^{1}$\thanks{\, Equal Contribution.}, $\,$ Zhen Han$^{2,3*}$, $\,$  Zhou Su$^{1}$, $\,$  Jiliang Li$^{1}$, $\,$  Volker Tresp$^{2,3}$, $\,$  Yuyi Wang$^{1,4}$ \\
$^{1}$School of Cyber and Engineering, Xi'an Jiaotong University\\
$^{2}$Institute of Informatics, LMU Munich $\;$  
$^{3}$Corporate Technology, Siemens AG\\
$^{4}$CRRC Zhuzhou Institute Co., Ltd.\\
\texttt{guojin0080@163.com, hanzhen02111@163.com}\\}
  
\begin{document}
\maketitle
\begin{abstract}
There has been an increasing interest in modeling continuous-time dynamics of temporal graph data.
Previous methods encode time-evolving relational information into a low-dimensional representation by specifying discrete layers of neural networks, while real-world dynamic graphs often vary continuously over time. Hence, we propose Continuous Temporal Graph Networks (CTGNs) to capture continuous dynamics of temporal graph data. We use both the link starting timestamps and link duration as evolving information to model continuous dynamics of nodes. 
The key idea is to use neural ordinary differential equations (ODE) to characterize the continuous dynamics of node representations over dynamic graphs. We parameterize ordinary differential equations using a novel graph neural network. The existing dynamic graph networks can be considered as a specific discretization of CTGNs. Experiment results on both transductive and inductive tasks demonstrate the effectiveness of our proposed approach over competitive baselines.
\end{abstract}

\section{Introduction}

Graph neural networks (GNNs)  have attracted growing interest in the past few years due to their universal applicability in various fields, \textit{e.g.}, social networks \cite{GnnForRecommend}  and natural language processing \cite{gnnNLPliu2021retrievalaugmented}. Graph neural networks (GNNs) learn a lower-dimensional representation for a node in a vector space by aggregating the information from its neighbors using discrete hidden layers. Then the embedding can be used for downstream tasks such as node classification \cite{NodeClassificition1}, link prediction \cite{GNN_LINK,Link2}, and knowledge completion \cite{GNN_KCompletion}.

\begin{figure}
    \centering
    \includegraphics[width=0.5\textwidth]{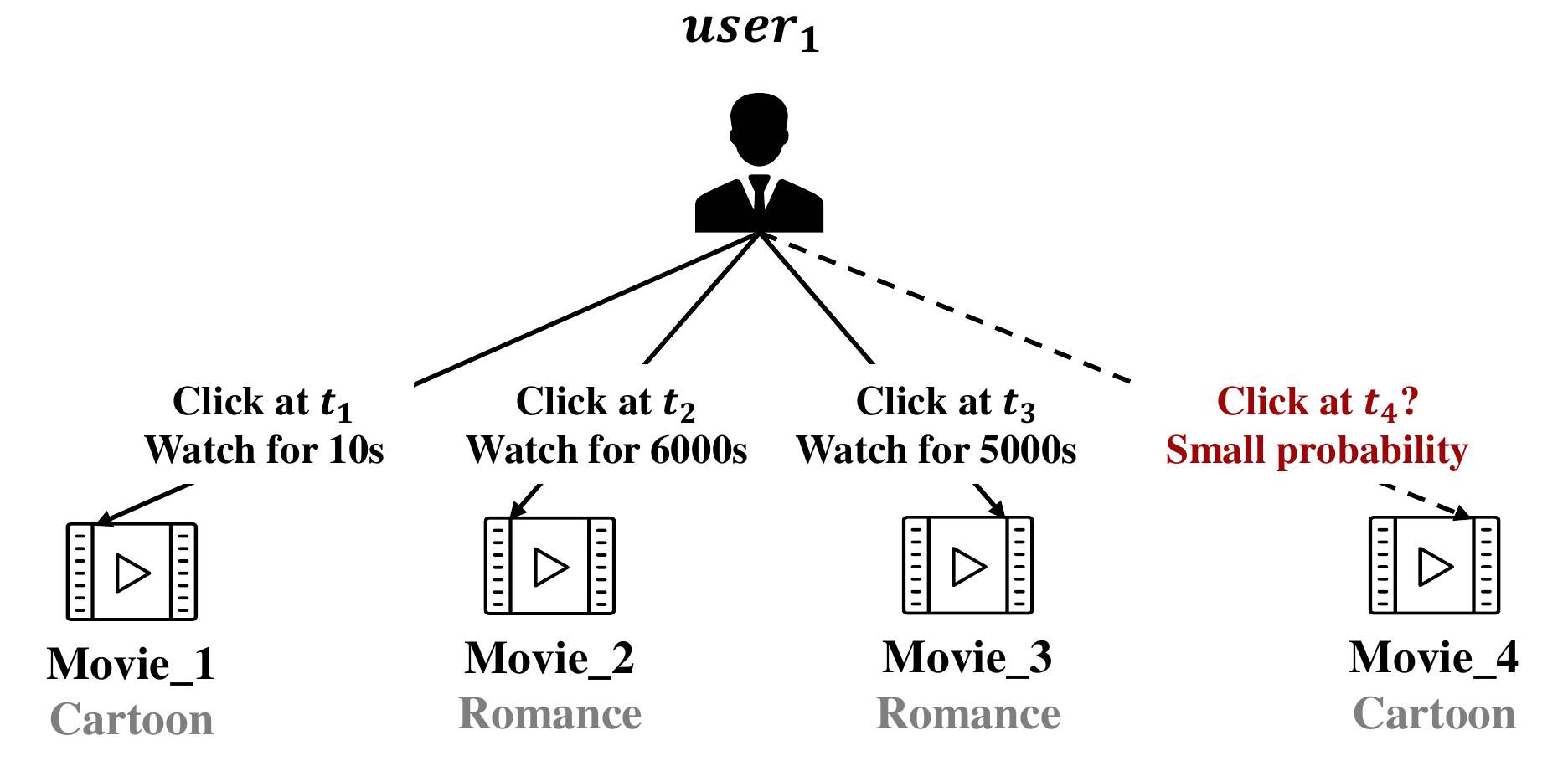} 
    \caption{The importance of link duration. Consider the behavior of a user watching movies. There are two types of nodes in the graph: $user$ nodes and $item$ nodes. Given the user's historical behavior,  the predicted target is (\textit{user$_1$}, \textit{don't\_click}, \textit{Movie\_4}). If we ignore the link duration information, $user_1$  seems interested in cartoon movies because he clicked on it at timestamp $t_1$. But $user_1$ only watched the \textit{Movie\_ 1} for 10s. The link duration indicated that although the user clicked, he was not interested.
     } 
    \label{fig:1}  
\end{figure}

Most graph neural networks only accept static graphs as input, although real-life graphs of interactions, such as user-item interactions, often change over time. Learning the node representation on dynamic graphs is a very challenging task. Dynamic graph methods can be divided into discrete-time dynamic graph (DTDG) models and continuous-time dynamic graph (CTDG) models. More recently,  an increasing interest in CTDG-based graph representation learning algorithms can be observed \cite{TGAT,Dyrep,Wikipedia,TGN,2020APAN,DBLP:journals/corr/abs-2101-05151}. 


 \begin{figure*}[htbp]
\centering

{\includegraphics[height=4.5cm]{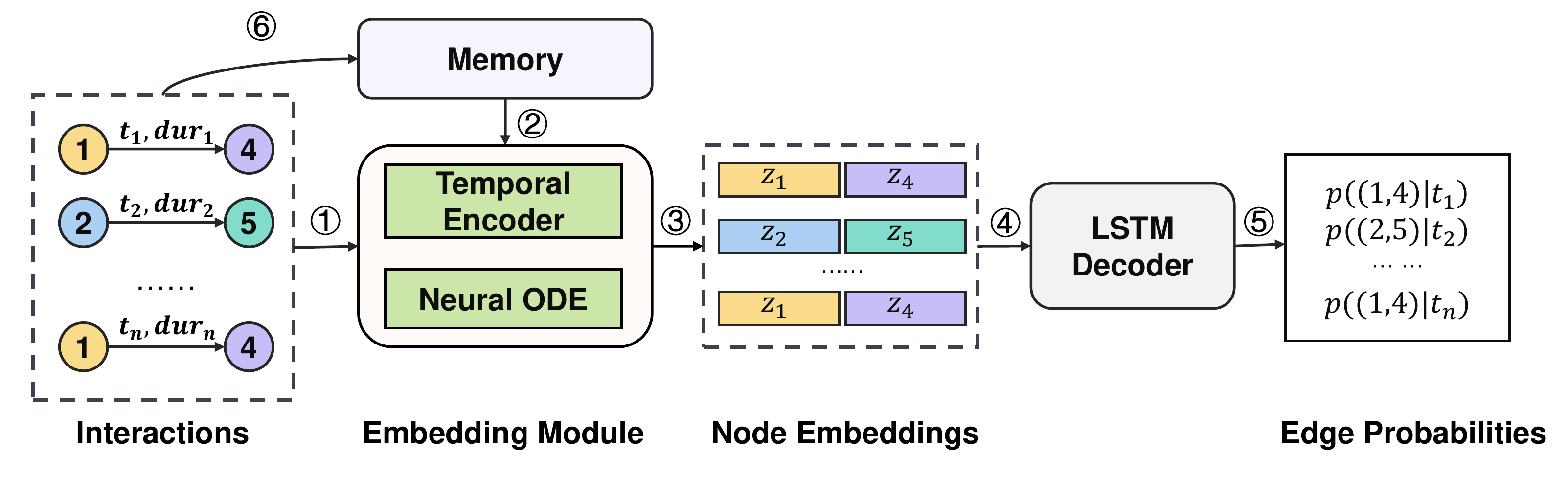}} 

\caption{Overview of our Continuous Temporal Graph network.  
 } 
\label{fig1}  
\end{figure*}
Although the above continuous-time dynamic methods have achieved impressive results, they still have limitations. The majority of research \cite{TGN,2020APAN,TGAT,Dyrep,Wikipedia} pays attention to the contact sequence dynamic graphs, in which the links are permanent, and no link duration is provided (\textit{e.g.}, email networks and citation networks).
However,  most real-life networks are event-based dynamic graphs in which the interactions between source nodes and destination nodes are not permanent (\textit{e.g.}, employment networks and proximity networks).  The event-based dynamic graph includes the time at which the link appeared and the duration of the link.
Link duration reflects the degree of association between the two nodes, \textit{e.g.}, user $i$ browses item $ j$ for 2 seconds and $k$ for 20 seconds. It means that the user's interest in the two items $j,k$  is different.  
Ignoring the link duration information can reduce the link prediction ability and even result in questionable inference.
Thus, it is crucial to consider the influence of link duration on node relationship prediction \cite{GNN_LINK,Link2} and knowledge completion \cite{GNN_KCompletion}.

The existing GNN-based methods \cite{AProposalonMachineLearningviaDynamicalSystems,2019Graph} that learn the node representation over dynamic graphs can be considered discrete dynamical systems. 
Chen \textit{et al.}~\shortcite{NODE} demonstrate that the continuous dynamical systems are more efficient for modeling continuous-time dynamic data.
The discrete networks can roughly be regarded as continuous networks by stacking enough layers. However, Onno and Suzuki~\shortcite{2019Graph} point out that graph neural networks (GNNs) exponentially lose expressive power for downstream tasks, which will lead to over-smoothing problems as we add more hidden layers. 
Therefore, designing effective continuous Graph Neural Networks to model continuous-time dynamics of node representation on dynamic graphs is critical.
To this end, many continuous graph neural networks \cite{NODE,CGNN} have been proposed recently. Although  those mentioned above continuous dynamic neural networks are more efficient to model the graph data, few approaches have been proposed for dealing with dynamic graphs using continuous-time dynamic neural networks.

This paper proposes a general framework of continuous temporal graph networks (CTGNs) to model continuous-time representations for dynamic graph-structured data. We combine Ordinary Differential Equation Systems (ODEs) and 
graphs methods. Instead of specifying discrete hidden layers, we integrate neural layers over continuous time. Figure \ref{fig1} illustrates the workflow of the proposed CTGN method. 
There is an interaction between two nodes. 
First, a novel temporal graph network (TGN)  is applied as the encoder to learn the latent states using the updated memory. Then, the neural ODE module is used to model the node’s continuous-time representation. Considering that the link duration reflects the degree of association between the two nodes, we use the link duration as the integration variable to control the weights of different interactions. After that, we use the LSTM \cite{LSTM} as the decoder to compute the probability of interaction between the two given nodes. Finally, the memory is updated as the input of the encoder. Memory is a compressed representation of the historical behavior of
all nodes defined in Section \ref{3.1}.
Experimental results on five real-world datasets of link prediction demonstrate the effectiveness of the proposed method over the state-of-art baselines. The main contributions of this paper are:
\begin{itemize}
\item We present a novel Continuous Temporal Graph Network (CTGN)  inspired by the neural ODE method. 
\item CTGNs pay attention to the event-based dynamic graph. CTGNs update the node’s representation with both the valid discrete timestamps when the link appears and the link duration between two linked nodes as evolving information.
\item We show that our model can outperform existing state-of-the-art methods on both transductive and inductive tasks.
\end{itemize}

\section{Background}

\subsection{ Dynamic Graph Methods}
\label{2.1}
The existing dynamic graph representation learning methods can be divided into two categories, discrete-time dynamic graphs and continuous-time dynamic graphs.

\textbf{Discrete-time dynamic graphs }(DTDGs) are a sequence of snapshots at different time intervals. 
\begin{equation}
DG = \{ G^1,G^2,...,G^T \}\,,\label{eq}
\end{equation}
where $T$ is the number of snapshots.
Current dynamic graph
methods \cite{2020EPNE,Know-Evolve,2019DynGraphGAN} have been mostly designed for discrete-time dynamic graphs (DTDGs).

\textbf{Continuous-time dynamic graphs} (CTDGs) can be viewed as a set of observations/events \cite{2019ASurvey}, and the network evolution information is retained. There are only a few works on CTDG. But recently, more attention has been paid to continuous-time graphs. All three representations of CTDG are described in more detail below.

\begin{enumerate}
\item \textbf{The contact sequence dynamic graph} is the simplest representation form of CTDG.

\begin{equation}
CS = (u_i,v_i,t_i)\,,\label{eq}
\end{equation}
where $u$ is the source node,  $v$ is the destination node, and  $t$ is the timestamp when the link appears.  
			In the contact sequence dynamic graph, the link is permanent (\textit{e.g}., citation networks) or instantaneous (\textit{e.g.}, email networks). Therefore, this graph has no link duration. 

There has been a lot of research on contact sequence dynamic graphs.
Trivedi \textit{et al.}~\shortcite{Dyrep} learn the representation of node $i$ by aggregating the node destination’s neighborhood information and updating the embedding for the node using a recurrent architecture after an interaction involving node $i$. Kumar \textit{et al.}~\shortcite{Wikipedia}
 employ two recurrent neural networks to update the embedding of a user and an item at every interaction. TGAT \cite{TGAT} proposes a novel functional time encoding method and uses self-attention to inductive representation learning on temporal graphs. Wang \textit{et al.}~\shortcite{2020APAN} propose the asynchronous propagation attention network (APAN) for real-time temporal graph embedding.
\item \textbf{The event-based dynamic graph} consists of the node pairs $(u,v)$,  the edge appears timestamp $t$ and the link duration $\Delta t$ . Link duration indicates how long the edge lasts until it disappears. 
\begin{equation}
EB = (u_i,v_i,t_i,\Delta t_i)\,.\label{eq}
\end{equation} \citet{TGN}  proposes a generic inductive framework operating on contact sequence dynamic graphs by adding a memory module on TGAT \cite{TGAT}. TGN can also operate on the event-based dynamic graph by simply replacing the timestamp $t$  with link duration $\Delta t$ in the memory module. 
\item \textbf{The streams graph} can be viewed as a particular case of the event-based dynamic graph. The streams graph includes the edge label $\delta $, which indicates edge removal or edge addition. 
\begin{equation}
GS = (u_i,v_i,t_i,\delta _i), \delta _i \in [-1,1] \,.\label{eq}
\end{equation}
TGN \cite{TGN} converts the streams graph into an event-based graph for processing. According to the edge label,  the event can be reorganized as $ (u_i,v_i,t',t)$, which was created at time $t'$ and deleted at time $t$, then two messages can be computed for the source and target nodes.

\end{enumerate}

The existing CTDG methods model discrete dynamics representations of continuous-time graph data with multiple discrete propagation layers. Our proposed method focuses on the event-based temporal graph and updates the node’s representation with both the timestamps and the link duration between the two nodes. CTGN also supports contact sequence dynamic graph. The model details will be slightly different from event-based dynamic graph. We will clarify this point in Chapter 3.
\subsection{Continuous-time Dynamical Systems }
Continuous-time dynamical systems mean that the system's behavior  changes with  time development  in the continuous-time domain. There have been related works that view data as a continuous object in artificial intelligence, \textit{e.g.}, pictures \cite{NODE} and static graphs \cite{CGNN,GDE}. The continuous-time dynamic graph (CTDG) we introduced in Section \ref{2.1} is also a continuous-time dynamical system in which nodes' state changes over time. Therefore, it is necessary to model the continuous dynamical system of CTDG data. To the best of our knowledge, our CTGN is the first approach that learn continuous-time dynamics on CTDG.
\subsection{Neural Ordinary Differential Equations and Continuous Graph Neural Networks}
Considering a residual network:
\begin{equation}
\textbf h_{t+1} = \textbf h_t + f(\textbf h_t,  \theta _t) 
\label{eq}
\end{equation}

A theoretical method to improve the performance of discrete networks is to stack more neural layers and take smaller steps \cite{NODE}. However, this scheme is not feasible because of the limited computer resources and over-fitting problems. Oono and Suzuki~\shortcite{2019Graph} point out that Graph Neural Networks (GNNs) exponentially lose expressive power for downstream tasks when adding more hidden layers because of over-smoothness problems.

 Inspired by residual network and ordinary difference, neural ordinary difference is proposed to solve this problem. Neural ODE models continuous-time dynamical systems by parameterizing the hidden state's derivative   using a neural network.

\begin{equation}
\frac{d_\textbf{z}}{d_t} =f(\textbf{z},t),\textbf{z}(0)=\textbf{x}, \label{eq}
\end{equation}
NeuralODE can be regarded as a discrete network with an infinitesimal learning rate and infinite layers. 
Weinanl~\shortcite{AProposalonMachineLearningviaDynamicalSystems} 
proposes the idea of using continuous dynamical systems to model hidden layers.
Chen\textit{ et al.}~\shortcite{NODE} introduce neural ODE, a continuous-depth model by parameterization the derivative of the hidden state using a neural network.
Neural ODE only focuses on unstructured data.
Xhonneux \textit{et al}.~\shortcite{CGNN}  apply continuous dynamical methods to static graph-structured data. They propose Continuous Graph Neural Networks (CGNNs), which solve the over-smoothing caused by stacking more layers and improve the performance of GNNs. 
Zang and Wang~\shortcite{NDCN} learn continuous-time dynamics on complex networks. 
However, continuous graph neural networks (CGNN) can only deal with static data.

\section{The Proposed Method: CTGN}
In this section, we introduce our proposed approach. The key idea of the CTGN is to build continuous-time hidden layers which can learn continuous informative node representations over event-based dynamic graphs. To characterize the continuous dynamics of node representation, we use ordinary differential equations (ODEs) parameterized by a neural network, which is a continuous function of time. We study both transductive and inductive settings. 
In the transductive task, we predict future links of the nodes observed during the training phase. In the inductive tasks, we predict future links of the nodes never seen before.
We first employ a temporal graph attention layer \cite{TGAT} to project each node into a latent space based on its features and neighbors. And then, an ODE module is designed to define the continuous dynamics on the node's latent representation $\textbf{h}_i (t)$. 


 \begin{table*}[h]

    \centering
    \setlength{\tabcolsep}{0mm}
 
    {
    \begin{tabular} {cccccc}
    \hline
  
    &\multicolumn{3}{c}{Event-based dynamic graph}&
    \multicolumn{2}{c}{Contact sequence dynamic graph} \\
    \cmidrule(r){2-4}  \cmidrule(r){5-6} 
    
    &NetFlix&Mooc&Lastfm&Wikipedia&Reddit\\
    \hline
    Nodes&18672&13374&7353&9227&10984 \\
    Edges&163417&131660&73358&157474&672447 \\
    Chronological Split & 70\%-15\%-15\% & 70\%-15\%-15\% & 70\%-15\%-15\%  & 70\%-15\%-15\%&70\%-15\%-15\%\\
    Unseen nodes&10\%&10\%&10\%&10\%&10\% \\
    Timespan&2 years&2 years&2 years&30 days&30 days\\
    \hline
    \end{tabular} }
  	\caption{Statistics of the datasets used in our experiments.}
  \label{dataset}
  \end{table*}

\subsection{Temporal Graph Network }
\label{3.1}
\textbf{Memory Passing.} Memory $\textbf{s}_i(t)$ is used to record the historical information of each node $i$ the model has seen so far. It is a compressed representation of the historical behavior of all nodes. Memory $\textbf{s}_i(t)$ is updated when there is an interaction involving node $i$.
At the end of each batch, we firstly compute memory $\textbf{s}_i(t)$ using the last time message $\textbf{m}_i(t^-)$ and memory $\textbf{s}_i(t^- )$: 
\begin{equation}
\textbf{s}_i (t)=mem(\textbf{m}_i (t),\textbf{s}_i (t^- ))\,.\label{eq}
\end{equation}
Here, $mem(\cdot)$ is a learnable memory update function. In all experiments, we choose the memory function as GRU. $\textbf{s}_i(0)$ is initialized as a zero vector.
At the end of each batch, the message $\textbf{m}_i(t)$ for the node  can be updated to compute $i$'s memory:
\begin{equation}
\begin{aligned}
\textbf{m}_i (t)=msg_s (\textbf{s}_i (t^- )||\textbf{s}_j (t^- )||\Delta t||\textbf{e}_{ij} (t))\,,
\\
\textbf{m}_j (t)=msg_s (\textbf{s}_j (t^- )  || \textbf{s}_i (t^- )||\Delta t||\textbf{\textbf{e}}_{ij} (t))\,.
\end{aligned}
\end{equation}
Here $||$ is the concatenation operator, $\Delta t$ is the link duration between node $i$ and $j$, . In the contact sequence dynamic graph, the link duration property is not available. We use $(t-t^-)$ as $\Delta t$. There may be multiple events $\textbf{e}_{i1} (t_1 ),…,\textbf{e}_{iN} (t_N ) $ involving the same node $i$ in the same batch. In the experiment, we only use the latest interaction $\textbf{e}_{iN} (t_N )$ to compute $i$’s message. $msg(\cdot)$ is a learnable function, and we use an RNN network in our experiment:

\textbf{Multi-head Attention.} Given an observed event $p=(i,j,t,\Delta t)$, we can compute the node latent representation respectively for $i$ and $j$ using:

\begin{equation}
\textbf{H}^{(l) } (t)=Attn^{(l) } (\textbf{Q}^{(l) } (t),\textbf{K}^{(l) } (t),\textbf{V}^{(l) } (t))\ ,\label{eq}
\end{equation}\begin{equation}
Attn (\textbf{Q},\textbf{K},\textbf{V}) = softmax(\frac{\textbf{Q}\textbf{K}^T}{\sqrt{d_k}})\textbf{V}\ ,
\label{eq}
\end{equation}
where \textbf{Q} , \textbf{K} , \textbf{V} denote the 'querys', 'keys', 'values', respectively. $\textbf{H}^{(l) } =[\textbf{h}_1^{(l) },...,\textbf{h}_i^{(l) }]$ are  the embedding of
the graph nodes of $l$-th layers.
The multi-head attention layer compute the node $i$'s representation by aggregating it's N-hop neighbors. 
\begin{equation}
\begin{aligned}
\textbf{Q}^{(l) } (t)=(\textbf{H}^{(l-1) } (t) \ || \ \phi(0))\textbf{W}_Q \,, \label{eq}
\end{aligned}
\end{equation}
\begin{equation}
\begin{aligned}
\textbf{K}^{(l) } (t)= \textbf{C}^{(l)}(t)\textbf{W}_K\,,
\label{eq}
\end{aligned}
\end{equation}
\begin{equation}
\begin{aligned}
\textbf{V}^{(l) } (t) \ = \textbf{C}^{(l)}(t)\textbf{W}_V\,,
\label{eq}
\end{aligned}
\end{equation}\begin{equation}
\begin{aligned}
\textbf{C}^{(l) } (t)=[\textbf{H}_1^{(l-1) } (t) \ || \ \textbf{E}_{1} (t_1 ) \ || \  \phi(t-t_1 ),\\…, 
\textbf{H}_N^{(l-1) } (t) \ || \ \textbf{E}_{N} (t_N ) \ || \  \phi(t-t_N ) ] \,.\label{eq}
\end{aligned}
\end{equation}

Here $ {\phi(\cdot)}$  represents a generic time encoder \cite{TGAT}. $\textbf{W}_Q,\textbf{W}_K,\textbf{W}_V \in \mathbbm{ R}^{d_k\times d_k}$ are the projection matrices used to generate attention embedding.
We define keys and values as the neighbor information. $ \textbf{h}_i^{(0) } (t)=\textbf{s}_i (t)+\textbf{v}_i$, $ \textbf{s}_i (t) $ is node $i$’s memory which saves the history information for the node. $\textbf{E}_n(t) = [e_{1n}(t),...,e_{in}(t)]$, $e_{in}(t)$ is edge features between node $i$ and it's $n$-hop neighbor at time $t$. 
Temporal graph network is a discrete method that can be thought of as a discretization of the continuous dynamical systems.

\begin{table*}[htb]
\centering
\setlength{\tabcolsep}{1.5mm}{
\begin{tabular}{ccccccccc}
\hline
&\multicolumn{2}{c}{NetFlix}&\multicolumn{2}{c}{Mooc} &\multicolumn{2}{c}{Lastfm}\\
\cmidrule(r){2-3}  \cmidrule(r){4-5} \cmidrule(r){6-7}
&Transductive&Inductive&Transductive&Inductive&Transductive&Inductive&\\
\hline
GAT*&96.45 $\pm$ 0.2&92.09 $\pm$0.6&83.33 $\pm$10&77.39 $\pm$10&76.77 $\pm$0.5&62.81 $\pm$0.6&\\
GraphSAGE*&95.14 $\pm$0.6&89.84 $\pm$1.7&82.01 $\pm$2.4&78.36 $\pm$2.2&77.41 $\pm$0.6&62.57 $\pm$0.3&\\
CGNN*&91.82 $\pm$0.2&†&96.88 $\pm$0.2&†&74.93 $\pm$10&† &\\
NDCN*&90.70 $\pm$0.9&†&96.07 $\pm$0.1&†&82.09 $\pm$1.4&† &\\
DyRep&99.07 $\pm$0.1&97.36 $\pm$0.1&83.52 $\pm$6.5&68.96 $\pm$4.0&82.96 $\pm$0.3&68.06 $\pm$0.3\\
Jodie&99.20 $\pm$0.1&97.43 $\pm$0.1&93.12 $\pm$0.6&80.85 $\pm$1.2&84.41 $\pm$0.3&68.14 $\pm$0.5\\
TGAT&96.56 $\pm$0.2&93.04 $\pm$0.2&73.69 $\pm$1.3&68.76 $\pm$1.2&78.80 $\pm$0.8&64.19 $\pm$0.7 \\
TGN&99.05 $\pm$0.2&97.38 $\pm$0.4&97.76 $\pm$0.4&93.86 $\pm$0.9&87.05 $\pm$0.1&72.89 $\pm$0.1\\
APAN& 98.23$ \pm$1.7& † &93.64 $\pm$1.3 & † & 82.65 $\pm$0.1 & †\\
\hline
CTGN&\textbf{99.27 $\pm$0.1}&\textbf{97.84 $\pm$0.2}&\textbf{97.97 $\pm$0.4}&\textbf{94.89 $\pm$0.4}&\textbf{87.20 $\pm$0.1}&\textbf{74.05 $\pm$0.1}\\
\hline
\end{tabular}}

\caption{ Experiments on event-based datasets. Average Precision (\%) for future edge prediction task in transductive and inductive settings. {\textbf{First}} best performing method. *Static graph method. †Does not support inductive.}
\label{table_transductive}
\end{table*}

 \begin{table*}[tbp]
\centering
  {
\begin{tabular}{ccccccccc}
\hline
&\multicolumn{2}{c}{node classification}&\multicolumn{2}{c}{link prediction-tranductive}&\multicolumn{2}{c}{link prediction-inductive} \\
\cmidrule(r){2-3} \cmidrule(r){4-5} \cmidrule(r){6-7} 
&Wikipedia&Reddit&Wikipedia&Reddit&Wikipedia&Reddit\\
\hline

GAE*&74.85 $\pm$0.6&58.39 $\pm$0.5&91.44 $\pm$0.1&93.23 $\pm$0.3&†&†\\
VGAE*&73.67 $\pm$0.8&57.98 $\pm$0.6&91.34 $\pm$0.3&92.92 $\pm$0.2&†&†\\
GAT*&82.34 $\pm$0.8&64.52 $\pm$0.5&94.73 $\pm$0.2&97.33 $\pm$0.2&91.27 $\pm$0.4&95.37 $\pm$0.3\\
GraphSAGE*&82.42 $\pm$0.7&61.24 $\pm$0:6&93.56 $\pm$0.3 &97.65 $\pm$0.2&91.09 $\pm$0.3&96.27 $\pm$0.2\\
DyRep&84.59 $\pm$2.2&62.91 $\pm$2.4&94.59 $\pm$0.2&97.98 $\pm$0.1& 92.05 $\pm$0.3&95.68 $\pm$0.2\\
Jodie&84.84 $\pm$1.2&61.83 $\pm$2.7&94.62 $\pm$0.5&97.11 $\pm$0.3& 93.11 $\pm$0.4&94.36 $\pm$1.1\\
TGAT&83.69 $\pm$0.7&65.56 $\pm$0.7&95.34 $\pm$0.1 &98.12 $\pm$0.2&93.99 $\pm$0.3&96.62 $\pm$0.3\\
TGN&87.81 $\pm$0.3&67.06 $\pm$0.9&98.46 $\pm$0.1 &98.70 $\pm$0.1&97.81 $\pm$0.1&97.55 $\pm$0.1\\
APAN&\textbf{89.86 $\pm$0.3}& 65.34 $\pm$0.4 & 98.12 $\pm$0.2 &\textbf{ 99.22 $\pm$0.2 }& † & † \\
\hline
CTGN&88.01 $\pm$1.5&\textbf{68.38 $\pm$3.4}&\textbf{98.64 $\pm$0.1}&98.28 $\pm$0.2&\textbf{98.01 $\pm$0.1}&\textbf{98.05 $\pm$0.2}\\
\hline
\end{tabular}}
\caption{  Experiments on contact sequence datasets. ROC AUC (\%) for the dynamic node
classification task, Average Precision (\%) for link prediction task. *Static method, †Does not support inductive. }
\label{node_classifi}
\end{table*}
  
\subsection{Model Continuous Dynamics of Node Representation}
In order to characterize the continuous dynamics of node representations, instead of only specifying a discrete sequence of hidden layers, we parameterize the hidden layers using ordinary differential equations (ODEs), a continuous function of time.

\begin{equation}
\frac{d_\textbf{z}}{d_t} =f(\textbf{z},t),\textbf{z}(0)=\textbf{x}. \label{eq}
\end{equation}
Here, $\textbf{x}$ is an initial vector, $f$ is a learnable function, $t$ is a time interval and $\textbf{z}$ is a vector.

\begin{equation}
\textbf{z} (t)=\textbf{z} (0)+\int_{0}^{t}(f(t,z))d\tau.
\label{eq15}
\end{equation}
We can compute  the node's continuous-time dynamics representation by Equation \ref{eq15} at arbitrary time t $>$ 0.

Previous work \cite{NDCN,GDE} model continuous-time dynamics for data by setting integration variable $[0,t]$ as a hyper-parameter.
 Considering the influence of link duration on the interaction between two nodes, we choose the link duration as the integration variable, in our experiment $t = dur$. 

Link duration shows how long it was (in seconds) until that user terminated browsing. Link duration can reflect the user's interest in different items. Take link duration as an integer variable that can control the weights of different interactions.


We parameterize the derivative of the hidden state using a neural network that takes the latent state, computed by the temporal graph network mentioned in Section \ref{3.1} as input. 
\begin{equation}
\textbf{z}_i (t)=ODESolver(f(t,z),\textbf{h}_i(t), \Delta t _i).\label{eq}
\end{equation}
Here, $\textbf{h}_i (t)$ is a discrete latent state computed by temporal graph networks, $\Delta t_i$ is the link duration between source node $i$ and destination $j$. $f(t,z)$ is ODE function, we choose $f(t,z)$ as MLP.
A black-box ODE solver computes the final node continuous dynamics embedding $\textbf{z}_i (t)$. We utilize the \textit{torchdiffeq.odeint\_adjoint} PyTorch package to solve reverse-time ODE and backpropagate.

\subsection{Time Smoothness}
The time-encoding method \cite{TGAT} used in this paper is an effective method to map timestamp $t$ from the time domain to d-dim vector space. However, the learning process of each timestamp is independent of other timestamps. Independent learning of hyperplanes of adjacent time intervals may cause adjacent times to be farther apart in embedded space. Actually, adjacent states in the graph should be more similar. To avoid the problem mentioned above, we constrained the variation between hyperplanes at adjacent timestamps by minimizing the euclidean distance:
\begin{equation}
L_{smooth} (W)=\sum_{t=1}^{T-1}||w_{t+1}-w_t||_2 \,.\label{eq}
\end{equation}
 
\begin{figure*}
\centering
\subfigure[]
{\label{a}\includegraphics[width=0.24\linewidth]{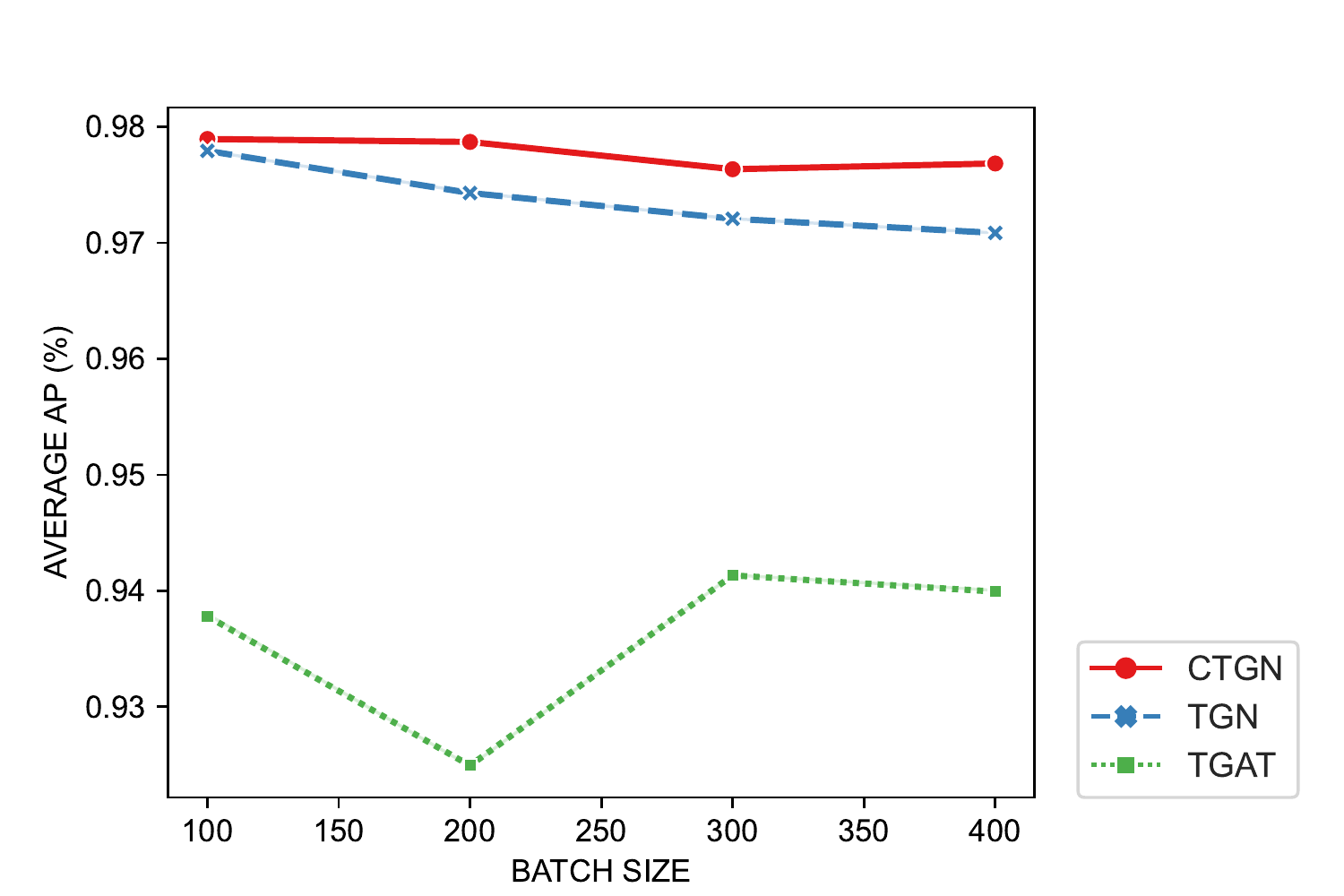}} 
\subfigure[]
{\label{b}\includegraphics[width=0.24\linewidth]{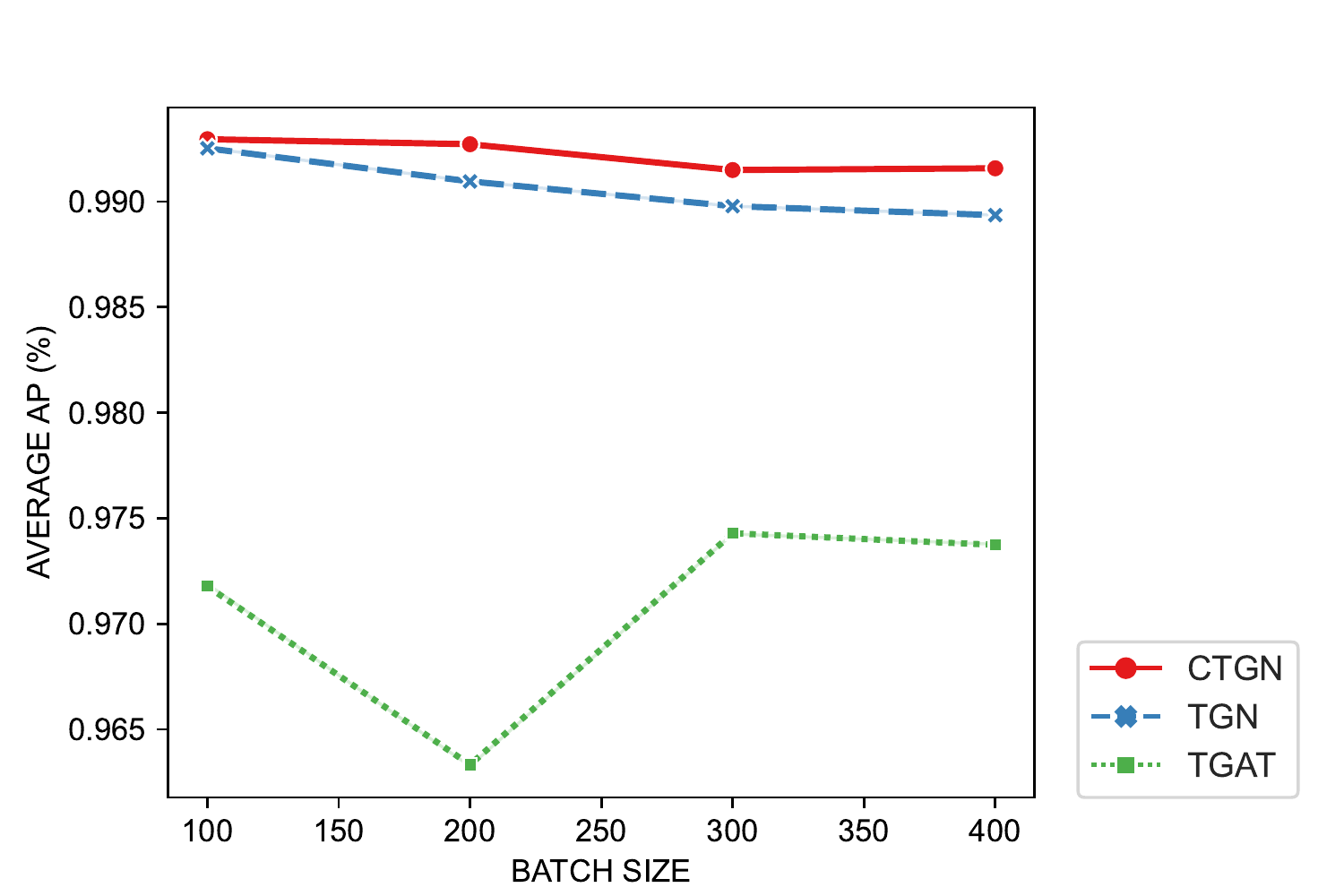}}
\subfigure[]
{\label{c}\includegraphics[width=0.24\linewidth]{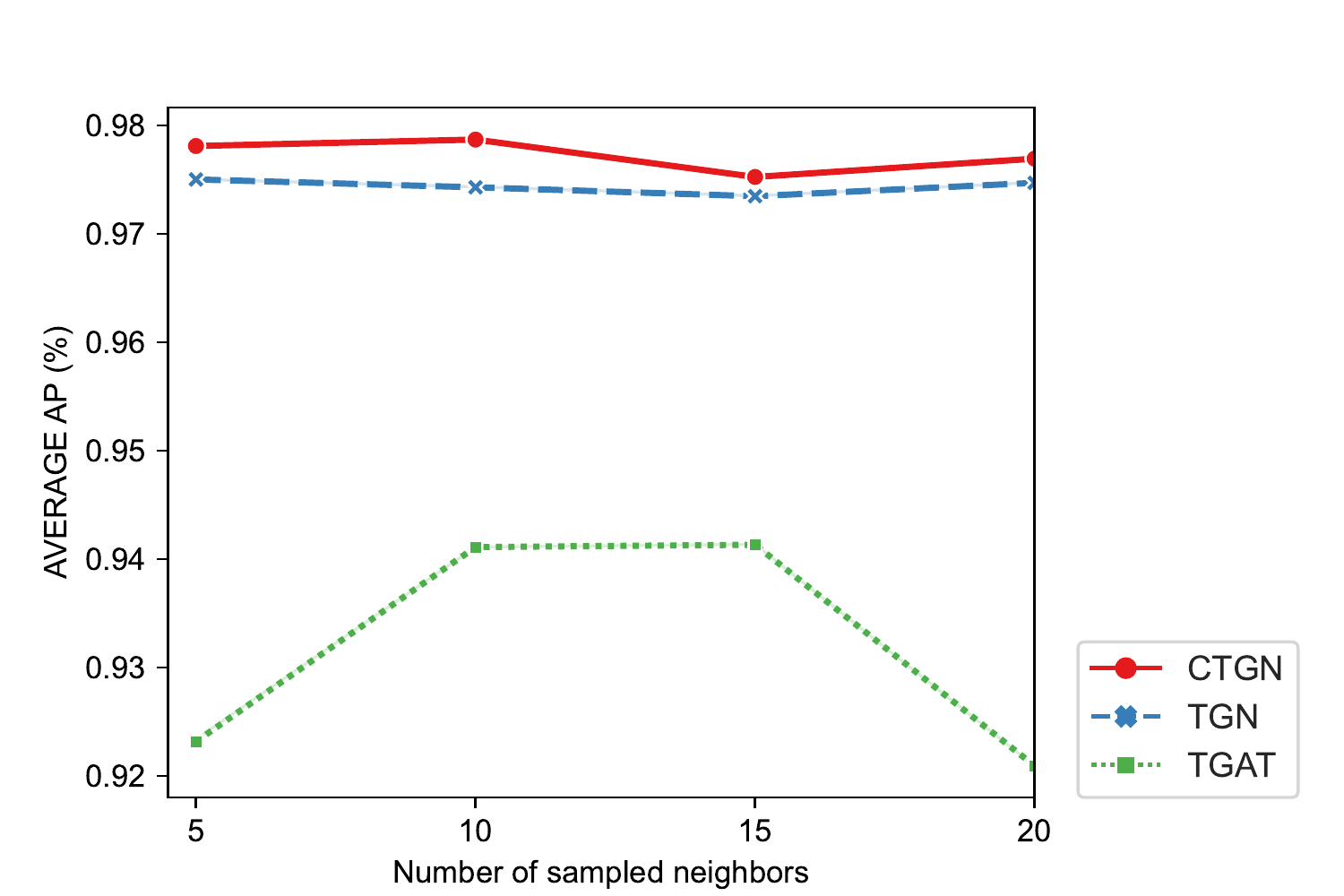}}
\subfigure[]
{\label{d}\includegraphics[width=0.24\linewidth]{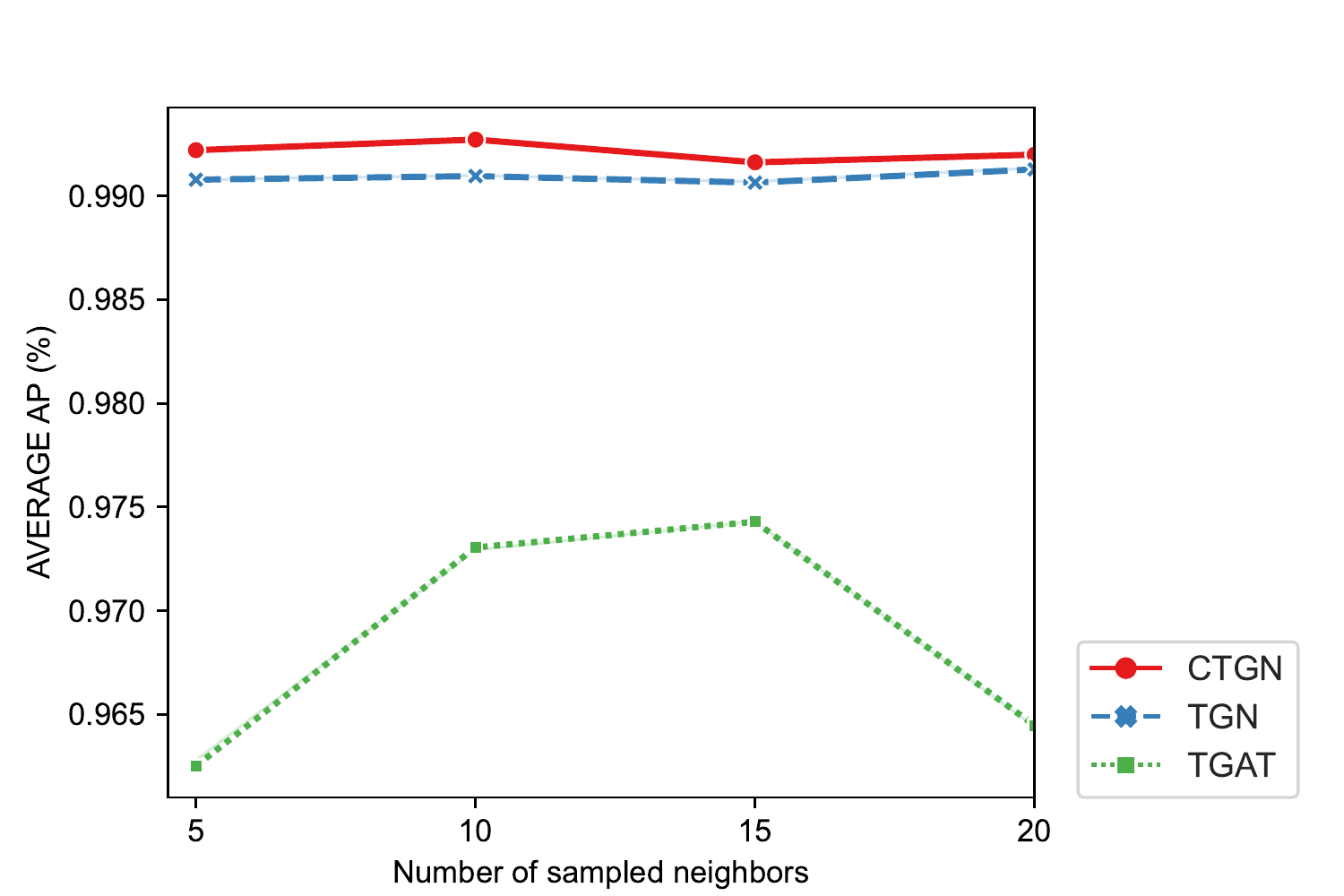}}
\caption{Ablation studies on the Netflix dataset for both the transductive and inductive setting of the link prediction task. \ref{a}  Sensitivity study result of batch size in inductive setting. \ref{b}  Sensitivity study result of batch size in transductive setting.  \ref{c} The relationship between number of sampled neighbors and the model performance in inductive setting. \ref{d} The relationship between number of sampled neighbors and the model performance in transductive setting. } 
\label{3}  
\end{figure*}
  
\subsection{Model Learning}
We use the link prediction loss function for training  CTGN:

\begin{equation}
loss=\alpha L_{smooth} (W)+L_{task}\,,\label{eq}
\end{equation}
where $\alpha$ is a tradeoff parameter, $ l_{task}$ is a loss function defined as the cross-entropy of the prediction and the ground truth. Our experiment found a parameter $\alpha$ of 0.002 for contact sequence dynamic graphs and 0.7 for event-based dynamic graphs.

\section{Experiment and Analysis}

In this section, we first introduce datasets, baselines and parameter settings. Then we compare our proposed method with other strong baselines and competing approaches for both the inductive and transductive tasks for two benchmarks contact  sequence dynamic graph datasets and three event-based dynamic graph datasets.

We study both transductive and inductive tasks. For event-based dynamic graphs, we learn link prediction tasks. For contact-sequence dynamic graphs, we learn dynamic node classification and link prediction tasks.

\subsection{Datasets}

We use five real-world datasets in our experiments, three event-based dynamic graphs: Netflix \footnote{\url{https://vodclickstream.com/}}, Mooc \cite{MOOC} and Lastfm \cite{LastfnDATA}, two contact sequence dynamic graphs: Wikipedia \cite{Wikipedia}, Reddit \cite{Wikipedia}.






The statistics of the datasets used in our experiments are described in detail in Table \ref{dataset}. 

\subsection{Baseline}

We compare our model with four CTDG methods: Jodie \cite{Wikipedia}, Dyrep \cite{Dyrep}, TGAT \cite{TGAT}, TGN  \cite{TGN}, APAN \cite{2020APAN}. And we also include four DTDG methods: GAE \cite{GAE_kipf2016variational}, VGAE \cite{GAE_kipf2016variational}, GAT \cite{GAT}, GraphSAGE \cite{GraphSAGE} as well as two state-of-the-art static graph neural ODE methods: CGNN \cite{CGNN},  NDCN \cite{NDCN}.







\subsection{Parameter Setup}
We set the batch size to 200 for training and patience to 5 for early stopping in all experiments. The node embedding dimension is 172. During training, we used 0.0001 as the learning rate for contact sequence dynamic graph datasets (Wikipedia and Reddit) and 0.00009 for event-based dynamic graph datasets (Netflix, Mooc, Lastfm). The weight of time smoothness loss $\alpha$ is set to 0.002 on Wikipedia , Reddit and 0.7 on Netflix, Mooc, Lastfm.
We choose the LSTM layer as the decoder for link prediction task and MLP for node classification task. We report mean and standard deviation across 10 runs. 
\subsection{Result}
To demonstrate the effectiveness of our proposed method, we compare CTGN with competitive baselines on five real-world event-based graph datasets. Table \ref{table_transductive} shows the results on link prediction tasks in both transductive and inductive settings for three event-based datasets. It is evident that our approach has achieved better results than the discrete dynamics graph neural networks on almost all datasets, especially in the inductive setting. 


 Table \ref{node_classifi} shows the dynamic node classification and link prediction results on two contact sequence-datasets. 
 CTGN has a solid ability to embed dynamic graphs. The conclusion can be obtained from the Table \ref{table_transductive} and Table \ref{node_classifi}.

Figure \ref{3} shows ablation studies on the Netflix dataset for both the transductive and inductive setting of the link prediction task. As we can see from Figure \ref{a} and \ref{b}, our model is not sensitive to batch size. When the training batch size is 100, CTGN has the same average precision as TGN. With the continuous increase of batch size, the performance of CTGN is more stable.

\section{Conclusion}
This paper introduces CTGN, a continuous temporal graph neural network for learning representation for event-based dynamic graphs. We build the connection between temporal graph networks and continuous dynamical systems inspired by neural ODE. Our framework allows the user to trade off speed for precision by selecting different learning rates and the weight of time smoothness loss parameters during training. We demonstrate on the link prediction task against competitive baselines that our model can outperform many existing state-of-the-art methods. 

\bibliography{anthology,custom}

\end{document}